\documentclass{article}
\usepackage{cite}
\usepackage{amsmath,amssymb,amsfonts}
\usepackage{algorithm}
\usepackage{algorithmic}
\usepackage{graphicx}
\usepackage{textcomp}
\usepackage{xcolor}
\usepackage{hyperref}
\usepackage{booktabs}
\usepackage{indentfirst}
\usepackage[a4paper, margin=1in]{geometry}

\begin{document}
\title{A Joint Prediction Method of Multi-Agent \\ to Reduce Collision Rate}

\author{Mingyi Wang, Hongqun Zou, Yifan Liu, You Wang, Guang Li}
\date{}

\maketitle

\begin{abstract}
Predicting future motions of road participants is an important task for driving autonomously. Most existing models excel at predicting the marginal trajectory of a single agent, but predicting joint trajectories for multiple agents that are consistent within a scene remains a challenge. Previous research has often focused on marginal predictions, but the importance of joint predictions has become increasingly apparent. Joint prediction aims to generate trajectories that are consistent across the entire scene. Our research builds upon the SIMPL baseline to explore methods for generating scene-consistent trajectories. We tested our algorithm on the Argoverse 2 dataset, and experimental results demonstrate that our approach can generate scene-consistent trajectories. Compared to the SIMPL baseline, our method significantly reduces the collision rate of joint trajectories within the scene.
\end{abstract}

Scene-Consistent Trajectories

\section{Introduction}

Accurate prediction of the motion of surrounding traffic participants is crucial for autonomous vehicles. Providing precise and timely predictions of intentions and trajectories is essential for downstream decision-making and planning modules, as it significantly enhances safety and the rationality of planned trajectories. Recent advances in deep learning have demonstrated great success in predicting accurate trajectories by learning from real-world driving examples\cite{b1}. 

However, many existing trajectory prediction models focus on generating marginal prediction samples for individual agents' future trajectories. How to effectively combine the different modal trajectories predicted for each agent remains a challenging problem. The reason is that the number of possible trajectory combinations grows exponentially as the number of traffic participants in the scene increases. This exponential growth is unacceptable for autonomous vehicles, which require quick responses. Meanwhile, marginal prediction overlooks the interactions between the predicted future trajectories of different agents, leading to potential collisions or overlaps in the predicted trajectories, which fails to adequately simulate the results of trajectory prediction in real-world scenarios.

\begin{figure}[htbp]
\centerline{\includegraphics[width=9.01cm,height=6.00cm]{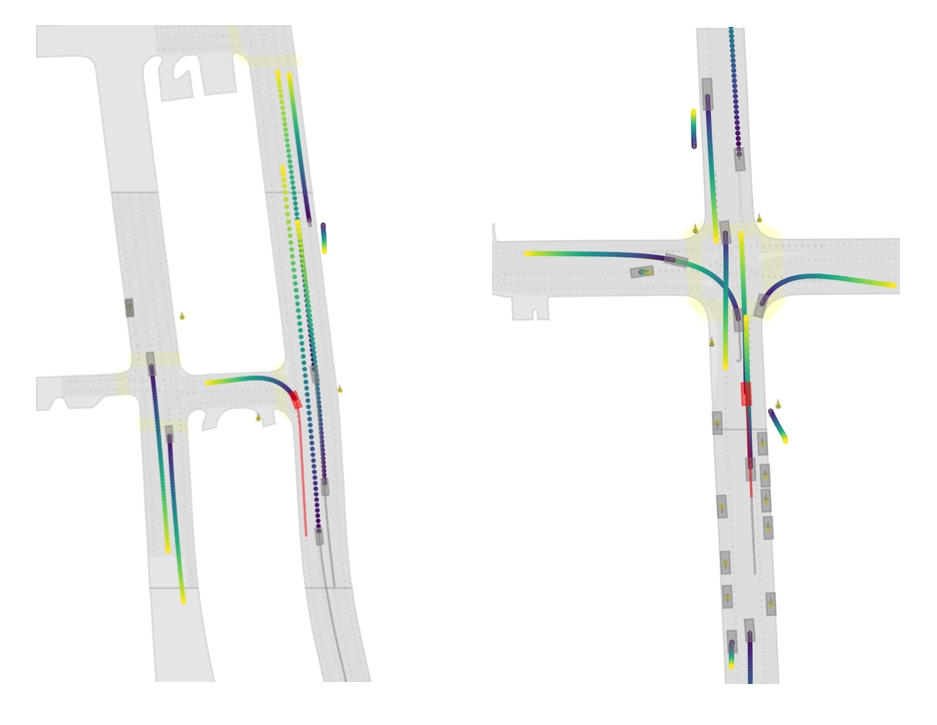}}
\caption{\textbf{Visualization of Predicted Trajectories for Agents in Scenarios.}
Our method can generate joint predictions for all agents in the scene simultaneously. The ego vehicle is shown in red, while other vehicles are displayed in gray. The predicted trajectories are visualized using gradient colors.}
\label{fig1}
\end{figure}

In many trajectory prediction models, the main goal is primarily on marginal prediction for a single target agent. This prediction typically accounts for both the agent's intentions and possible trajectories, offering multiple trajectory modes along with their associated probabilities\cite{b1}-\cite{b3}. This methodology has driven significant progress, resulting in the development of numerous high-performing models that have achieved notable success in both benchmark datasets and real-world applications.

However, this approach presents challenges when applied to joint prediction scenarios involving multiple agents. Specifically, when predicting the trajectories for two or more agents, selecting the highest probability paths for each agent independently often fails to capture the realistic dynamics of their interactions. The resulting trajectories may overlap or even collide, which indicates a failure to model the true interactions between agents in a shared environment.

A straightforward approach to generating joint future predictions involves considering the exponential number of combinations derived from the marginal predictions of individual agents. However, many of these combinations are inconsistent, particularly when agents have overlapping trajectories. Moreover, this method leads to an exponential increase in prediction complexity as the number of agents in a scenario grows\cite{b4}. A more natural approach is to allow the network to implicitly learn contextual features of trajectory prediction and to employ a scene-consistent method to generate joint trajectory predictions that are coherent across different scenarios\cite{b5}.

To address the issues of limited joint prediction methods and relative computational complexity mentioned above, our motivation is to propose a simple, scene-consistent joint prediction approach. In our research, we diverged from approaches that rely on recombining Factored Marginal Trajectories like M2I\cite{b4} or Conditional Prediction like SceneTransformer\cite{b6}. Instead, we employed an implicit scene learning method, designing a scene-level trajectory loss function. We utilized a Winner-Takes-All(WTA) backpropagation strategy to generate a set of scene predictions that are closest to the true scene trajectories. This approach enables the network to implicitly learn the distribution of scene-level trajectories.

Our method is capable of simultaneously generating coherent joint predictions for all agents concerning future trajectories. This represents a crucial step toward the joint optimization of Prediction and Planning in future research. The visualization results of our method are presented in Fig. \ref{fig1}.

\section{Related Works}

The trajectory prediction task typically consists of the following components: a context encoder, which encodes map information and the historical trajectories of agents; a context fusion model, which performs interactive modeling of the extracted context information, enabling the target agent to learn about the surrounding agents and map information; and a trajectory decoder, which decodes the high-dimensional latent variables obtained from the fusion model into predicted trajectories. During the trajectory decoding process, various approaches can be employed to generate scene-consistent joint predictions or marginal predictions.

\subsection{Context Encode}

In the field of trajectory prediction, context representation plays a crucial role. Most early approaches often represented the surrounding environment as a multi-channel bird’s-eye-view image\cite{b7}. However, recent research has increasingly adopted vectorized scene representations\cite{b8}, where locations and geometries are annotated using point sets or polylines with geographic coordinates. This approach enables more efficient retention of valuable information during network learning.

In scene-level trajectory prediction, two primary approaches for context representation involve coordinate system encoding. The first uses a shared coordinate system, such as one centered on the autonomous vehicle, which simplifies encoding and integrates easily with upstream perception results. However, it often suffers from reduced generalization, impacting network performance. The second approach, agent-centric encoding, normalizes the scene context relative to the target agent's current state, transforming coordinates based on the agent being predicted. This method generally performs well for marginal prediction but does not extend effectively to joint prediction scenarios.

In scene-level trajectory prediction, a commonly used coordinate system encoding method is the instance-centric coordinate system. When constructing scene-centric joint predictions, using target-centric encoding for all target agents in the scene requires repeating the normalization process and feature encoding for each agent. This not only leads to extensive redundant encoding but also imposes significant computational challenges. Therefore, the latest advancements in context encoding have shifted towards an instance-centric approach\cite{b9}.

\subsection{Context Fusion Methods}

In a frequently mentioned approach\cite{b8}, map elements are represented as polylines and sparse graphs, using raw coordinates to preserve spatial information. These features are further processed using Graph Neural Networks\cite{b2} or Transformers\cite{b8}, resulting in higher fidelity and efficiency.

After encoding the scene context in a vectorized format, road information and vehicle historical trajectories are typically modeled as tokens. The network then models these historical trajectories and map tokens, a phase that can be understood as feature fusion in traditional perception tasks. In this stage, most trajectory prediction frameworks opt to use Transformers for feature fusion, which is one of the key reasons for modeling input information as tokens. Common feature fusion strategies include using self-attention mechanisms to model multimodal trajectories and cross-attention mechanisms to extract scene information. For example, the symmetric fusion Transformer (SFT) used in SIMPL\cite{b1} employs the Actor feature as the query, while the context information fused from the actor feature, lane feature, and relative positional embedding (RPE) serves as the key and value in the attention mechanism. The fused Query is then used to provide information to the downstream Decoder. This method of context fusion is widely adopted in trajectory prediction and has become the mainstream approach for context integration.

\subsection{Joint Prediction Methods}

Predicting scene-compliant trajectories for multiple agents has always been a challenging task. Traditional approaches often rely on hand-crafted interaction models, such as social forces\cite{b10} and energy functions\cite{b11}. However, these methods require complex rules and struggle to capture intricate patterns in more complicated scenarios. As a result, an increasing number of researchers have turned to learning-based methods to achieve higher accuracy.

In contrast to Marginal Prediction, common approaches to joint prediction can be categorized into two main types. The first type explicitly models the influencer and reactor dynamics, where predictions are made in stages: first by predicting the labels and then performing conditional prediction based on the influencer's trajectory. A notable example of this approach is M2I\cite{b4}. The second type employs implicit methods to model the interaction process, directly outputting the agents' behavior across different scenarios. Notable examples of this approach include Autobots\cite{b5} and Scene Transformer\cite{b6}.

\section{Proposed Method}
\subsection{Problem Formulation}
The goal of trajectory prediction is to forecast potential future paths for target agents based on their past movements and relevant map data. In a driving environment involving \( N_a \) agents, including the autonomous vehicle (AV), the map information is represented by \( M \), while the observed trajectories are denoted as \( X = \{x_0, \ldots, x_{N_a}\} \). Each trajectory \( x_i \) represents the historical path taken by agent \( i \) over the last \( H \) time steps.

The objective is to estimate \( K \) possible future trajectories for each agent \( i \), accompanied by corresponding probability scores to represent the inherent multimodal nature of the predictions. The predicted trajectories are given by \( y_i = \{y_i^1, \ldots, y_i^K\} \), with associated probabilities \( \alpha_i = \{\alpha_i^1, \ldots, \alpha_i^K\} \). The multimodal marginal prediction can be formulated as a mixture distribution:
\begin{equation}
P(y_i \,|\, X, M) = \sum_{k=1}^{K} \alpha_i^{(k)} f(y_i^{(k)} \,|\, X, M) \label{eq1}
\end{equation}

\subsection{Model Structure}

\begin{figure}[htbp]
\centerline{\includegraphics[width=9.00cm,height=3.5cm]{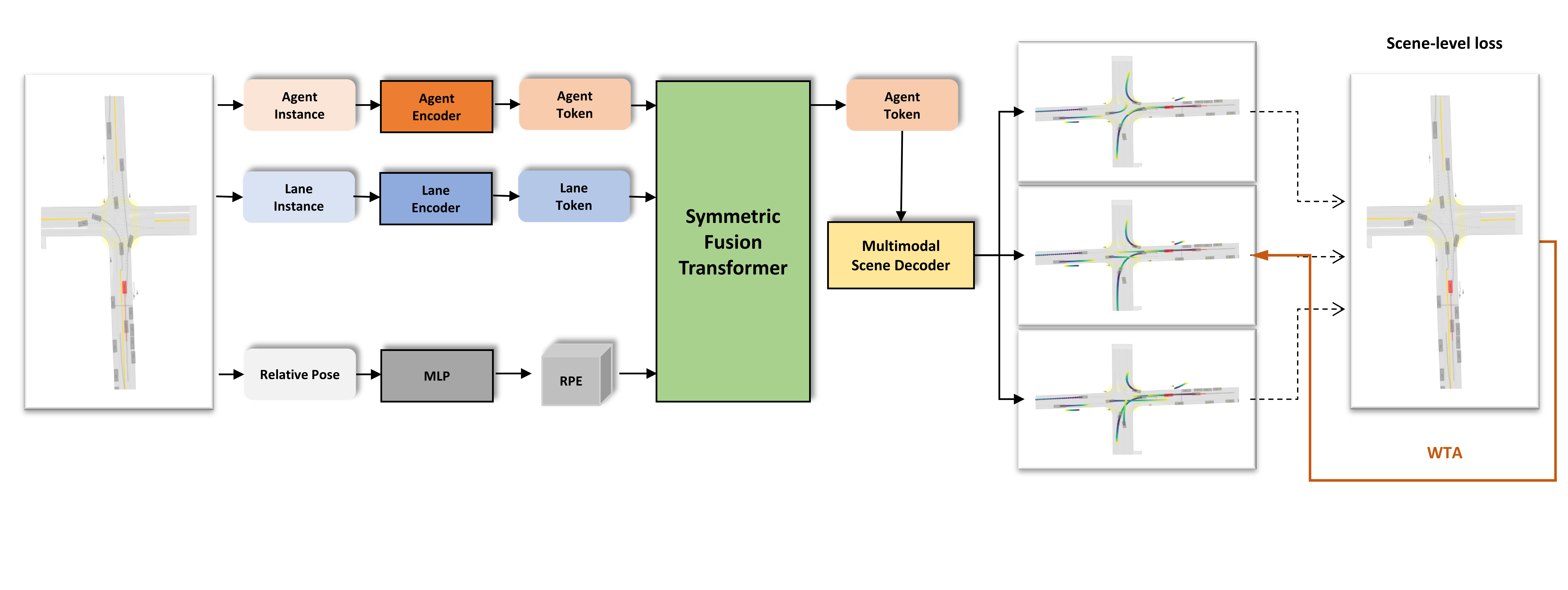}}
\caption{\textbf{Overall structure} Our model builds upon the SIMPL framework. Semantic instance features are encoded using simple encoders, relative positional embeddings, and a symmetric feature transformer. Subsequently, our proposed multimodal scene-consistent decoder and scene-consistent loss are used to train the model, generating scene-consistent results.}
\label{fig2}
\end{figure}

Our proposed method builds upon the SIMPL model\cite{b1} with several key improvements, which we detail in the following sections. First, we represent lane lines and vehicle historical trajectories using a vectorized scene representation. Then, we extract features for actors and the map using a feature encoder, and construct a relative position embedding (RPE) to capture the spatial relationships between them. In the fusion network, we employ the SFT module provided in SIMPL to model the interactions between vehicle information and lane line data. Finally, we use a multimodal scene decoder to generate joint trajectories that are consistent across the entire scene. The whole structuer is shown in Fig. \ref{fig2}.

Unlike ADC-centric and agent-centric approaches, instances in the scene (such as agents and lanes) can be represented by vectorized features under their local frames, along with the relative poses between them. Each instance has its own local coordinate system, so this modeling approach is also referred to as "instance-centric." In this framework, the vectorization of lane segments typically shows little variation, as most lane segments in a given scene are quite similar, which aligns with our common understanding. Therefore, the key factor in this approach is the relative positional embedding (RPE).

For two instances and their respective coordinate systems \( i \) and \( j \), the relationship between these two coordinate systems can be represented using three quantities: the heading difference \( \alpha_{i \rightarrow j} \), the relative azimuth \( \beta_{i \rightarrow j} \), and the distance \( \|d_{i \rightarrow j}\| \). To enable the network to better learn the relationship between the two instances, we use a five-dimensional vector to represent the relative positional embedding (RPE) between the two coordinate systems, given by \( r_{i \rightarrow j} = [\sin(\alpha_{i \rightarrow j}), \cos(\alpha_{i \rightarrow j}), \sin(\beta_{i \rightarrow j}), \cos(\beta_{i \rightarrow j}), \|d_{i \rightarrow j}\|] \). The calculation formula is as follows.
\begin{equation}
\sin(\alpha_{i \rightarrow j}) = \frac{\mathbf{v}_i \times \mathbf{v}_j}{\|\mathbf{v}_i\|\|\mathbf{v}_j\|}  \label{eq2}
\end{equation}
\begin{equation}
\cos(\alpha_{i \rightarrow j}) = \frac{\mathbf{v}_i \cdot \mathbf{v}_j}{\|\mathbf{v}_i\|\|\mathbf{v}_j\|} \label{eq3}
\end{equation}
\begin{equation}
\sin(\beta_{i \rightarrow j}) = \frac{\mathbf{d}_{i \rightarrow j} \times \mathbf{v}_j}{\|\mathbf{d}_{i \rightarrow j}\|\|\mathbf{v}_j\|} \label{eq4}
\end{equation}
\begin{equation}
\cos(\beta_{i \rightarrow j}) = \frac{\mathbf{d}_{i \rightarrow j} \cdot \mathbf{v}_j}{\|\mathbf{d}_{i \rightarrow j}\|\|\mathbf{v}_j\|} \label{eq5}
\end{equation}

For a scene with \( N \) instances, where \( N = N_{\text{agent}} + N_{\text{lane}} \), there is a five-dimensional relative positional embedding (RPE) between each pair of instances. Therefore, the shape of the positional embeddings for the entire scene is \( [N, N, 5] \).

We use the same Agent Encoder and Lane Encoder as in SIMPL, employing a 1D CNN-based network\cite{b2} to handle historical trajectories and a PointNet-based encoder\cite{b12} to extract static lane features. The encoded results are all transformed into \( D \)-dimensional vectors. Finally, we concatenate the encoded agent features with the encoded lane features, resulting in a tensor of dimensions \([N_{\text{agent}} + N_{\text{lane}}, D]\), where \( N_{\text{agent}} \) is the number of actors and \( N_{\text{lane}} \) is the number of map elements.

In the feature fusion stage, we use the Symmetric Fusion Transformer (SFT) to uniformly model the relationships between map elements and agents, as well as between agents themselves, similar to SIMPL. Leveraging the powerful modeling capabilities of the Transformer\cite{b14}, we can integrate agent tokens, which represent agent features, lane tokens, which represent map element features, and Relative Positional Embeddings (RPE) that capture the relative positions between elements within this framework, allowing it to autonomously learn the inherent correlations. The SFT module consists of several layers of standard Transformer, and uses agent tokens and lane tokens as queries, with the information fused with RPE through expansion and repetition as keys and values. For a specific token \( f_i \), its relationship with another token \( f_j \) and the relative position \( r'_{i \rightarrow j} \) is concatenated, then passed through an MLP network for feature fusion, resulting in the fused embedding feature \( e_{i \rightarrow j} \) as follows:
\begin{equation}
e_{i \rightarrow j} = \phi \left( \text{concat}\left(f_i, f_j, r'_{i \rightarrow j}\right) \right) \label{eq6}
\end{equation}

Next, cross-attention is employed to enable each token to be aware of the information from other tokens, which is the core mechanism of the SFT. The process using multihead attention (MHA) is shown in the following equation.
\begin{equation}
f'_i = \text{MHA}(\text{Query}: f_i, \text{Key}: E_i, \text{Value}: E_i) \label{eq7}
\end{equation}

Compared to graph neural networks, the Transformer can be understood as a fully connected graph network where each node is connected to every other node. Leveraging its powerful learning capabilities, it effectively learns the relationships between these nodes. Ultimately, after passing through multiple layers of the Transformer, the agent tokens are extracted as input for the trajectory decoder. By this stage, the agent tokens have already integrated information from the map and other agents, thus obtaining the context information.

\subsection{Multimodal Joint Trajectory Decoder}

After the symmetric global feature fusion, the updated actor tokens are gathered and sent to a multimodal motion decoder to generate predictions for all agents. Similar to using a multimodal trajectory decoder for each agent in the scene, we first generate \( K \) possible future trajectories. Following\cite{b2}, we predict \( K \) possible futures, and for each possible future, we use a scene-level MLP decoder to output the future trajectories of all agents in the scene.
After each decoder, the resulting output consists of the future trajectories of all agents in the scene. The final output trajectory has the dimensions \([N_{\text{agent}}, K, \text{PredLength}, 2]\), representing the \( x \) and \( y \) coordinates of all agents over the prediction length in \( K \) different predicted scenarios.

\subsection{Scene-consistent Loss}

Our method is capable of generating scene-consistent joint trajectory predictions and multimodal scenarios, with the core elements being the use of a scene consistency loss function and the Winner-Takes-All strategy\cite{b2}.
Unlike marginal prediction, where the regression loss is calculated by selecting the predicted trajectory closest to the ground truth for each individual agent, the scene consistency loss function computes the error at the scene level. Specifically, it calculates the difference between the predicted trajectory endpoints and the ground truth for all target agents in the scene, selecting the scenario with the smallest overall error.
During training, we employ WTA strategy to mitigate the issue of mode collapse and enable the network to learn multimodal trajectory information. Specifically, the regression loss is only computed for the scenario with the smallest error, where the error is the difference between the predicted trajectory endpoints of all agents in the scene and the ground truth. The predicted trajectories of all target agents in this scenario are then backpropagated using SmoothL1Loss.

The pseudocode is shown in Alg. \ref{alg1}.

\begin{algorithm}
\caption{Calculate Scene-Consistent Loss}
\label{alg1}
\begin{algorithmic}[1]
\STATE \texttt{Input \textbf{agent\_pred} shape [A, K, T, 2]}
\STATE \texttt{Input \textbf{agent\_gt} shape [A, T, 2]}
\STATE \COMMENT {Get the distance error between agent\_pred and agent\_gt shape [A, K, T]}
\STATE \texttt{dist\_error = norm(agent\_pred - agent\_gt.repeatdim(1), dim=-1)}

\STATE \COMMENT {Get the endpoint agent sum error per scene shape [K]}
\STATE \texttt{scene\_error = sum(dist\_error[:,:,-1], dim=0)}

\STATE \COMMENT{ Get the minimum endpoint error index per scene}
\STATE \texttt{scene\_index = argmin(scene\_error)}

\STATE \COMMENT{Get the joint loss}
\STATE \COMMENT{Only maintain gradient and calculate loss for all agents in the scene with min error}
\STATE \texttt{joint\_reg\_loss = SmoothL1Loss(agent\_pred
[:,scene\_index,:,:] - agent\_gt)}

\STATE \COMMENT{For winner-takes-all loss backwards}
\STATE \texttt{joint\_reg\_loss.backward()}
\end{algorithmic}
\end{algorithm}

We believe that this approach enables the model to distinguish between trajectories generated in different scenarios and naturally capture the diversity of scenes and trajectories, without forcing the model to explicitly model interactions. This results in scene-consistent joint predicted trajectories. The ${joint\_reg\_loss}$ is only the regression loss; to generate the scenario and its probability, a classification loss is also required. Following, our model is trained end-to-end, with the final loss function being a weighted average of the regression loss and classification loss, where \( \omega \in [0, 1] \) is the weight used to balance these components. We set \( \omega = 0.9 \) to emphasize the importance of the regression task.
\begin{equation}
\mathcal{L} = \omega \mathcal{L}_{\text{reg}} + (1 - \omega) \mathcal{L}_{\text{cls}} \label{eq8}
\end{equation}

\section{Experimental Results}
\subsection{Experiment Setup}
We evaluate the proposed method on the Argoverse 2 motion forecasting dataset\cite{b13}. This dataset consists of 200,000 sequences for training, 25,000 for validation, and 25,000 for testing. The sequences are sampled at 10 Hz, with a given history of 5 seconds and a prediction horizon of 6 seconds (\(H = 50\), \(T = 60\)). For map information, we use HD maps provided by Argoverse 2.

We set the latent vector dimension to \(D = 128\) for all tokens, and we use a stack of 4 SFT layers and 8 attention heads for symmetric global feature fusion. For the multimodal scene decoder, we follow common practices to set the number of modes to \(K = 6\). Our model is trained in an end-to-end manner with a batch size of 128 over 50 epochs. We use the Adam optimizer, setting the initial learning rate to $1 \times 10^{-3}$, which is gradually reduced to $1 \times 10^{-4}$ after 35 epochs.

\subsection{Metric}
In the field of trajectory prediction, evaluation metrics are typically derived from the comparison between predicted trajectories and ground truth. Common marginal prediction metrics include minimum average displacement error (minADE\(_k\)), minimum final displacement error (minFDE\(_k\)), and miss rate (MR\(_k\)). All these metrics evaluate the best-predicted trajectory for a single target agent among the \( K \) hypotheses against the ground truth. These metrics do not adequately evaluate the results of joint prediction.

Since these metrics are designed for marginal prediction, to adapt them for evaluating entire scenes, we need to calculate the average metrics across all agents that require prediction within the scene. The mean FDE associated with a predicted world, summarized across all scored actors within a scenario. The world with the lowest avgMinFDE is referred to as the "best" world. avgMinFDE\(_k\) is calculated as follows:

\begin{equation}
\text{avgMinFDE}_6 = \min_{k=1,2,\dots,K} \left( \frac{1}{N} \sum_{i=1}^{N} \left\|\mathbf{y}_T^i - \hat{\mathbf{y}}_T^{i,k}\right\| \right) \label{eq9}
\end{equation}

\( K \) represents the number of predicted "worlds", we assume \(K=6\) here, \( N \) represents the number of scored actors within a scenario, \( \mathbf{y}_T^i \) denotes the ground truth endpoint of the trajectory for the \( i \)-th actor, \( \hat{\mathbf{y}}_T^{i,k} \) denotes the predicted endpoint of the trajectory for the \( i \)-th actor in the \( k \)-th world, and \( \|\cdot\| \) represents the Euclidean distance.

In addition, to measure scene consistency, we use the collision rate to calculate the spatiotemporal collision rate among all agents in a given scenario. This metric physically represents whether the jointly predicted future trajectories will result in collisions, making it one of the most important metrics for assessing scene consistency and safety.

The pseudocode for determining whether a collision occurs in a given scenario is shown in Alg. \ref{alg2}.

\begin{algorithm}
\caption{Collision Detection Algorithm}
\label{alg2}
\begin{algorithmic}[1]
\STATE $n, timesteps \gets$ shape of $trajectories$
\FOR{$t = 1$ to $timesteps$}
    \FOR{$i = 1$ to $n$}
        \FOR{$j = i + 1$ to $n$}
            \STATE Calculate $distance \gets \|trajectories[i, t, :] - trajectories[j, t, :]\|$
            \IF{$distance < dist\_safe$}
                \STATE \textbf{return} True \COMMENT{Collision}
            \ENDIF
        \ENDFOR
    \ENDFOR
\ENDFOR
\STATE \textbf{return} False \COMMENT{No Collision}
\end{algorithmic}
\end{algorithm}

Based on this, we propose the CollisionRateK (CR\(_k\)) metric, which calculates whether collisions occur in the scenario with the smallest avgMinFDE\(_k\). If a collision occurs, the value is True; if no collision occurs, the value is False. The final CR\(_k\) is the average value obtained across the entire dataset being evaluated.

\subsection{Results}

To quantitatively analyze the results, we conducted experiments on the validation set of Argoverse 2, comparing our method with the SIMPL baseline. We selected the best-performing model of SIMPL on the Argoverse 2 dataset as the baseline for marginal prediction, and we compared four metrics: avgMinADE\(_k\), avgMinFDE\(_k\), avgMR\(_k\), and avgCR\(_k\). We believe these four metrics can effectively reflect the performance of trajectory prediction models, and avgCR\(_k\), as a safety metric for vehicle trajectory predictions, can measure the model's performance in scene consistency. The lower the collision rate, the better the model's scene consistency in joint predictions, and thus, the better the overall model performance.

\begin{figure}[htbp]
\centerline{\includegraphics[width=8.82cm,height=5.88cm]{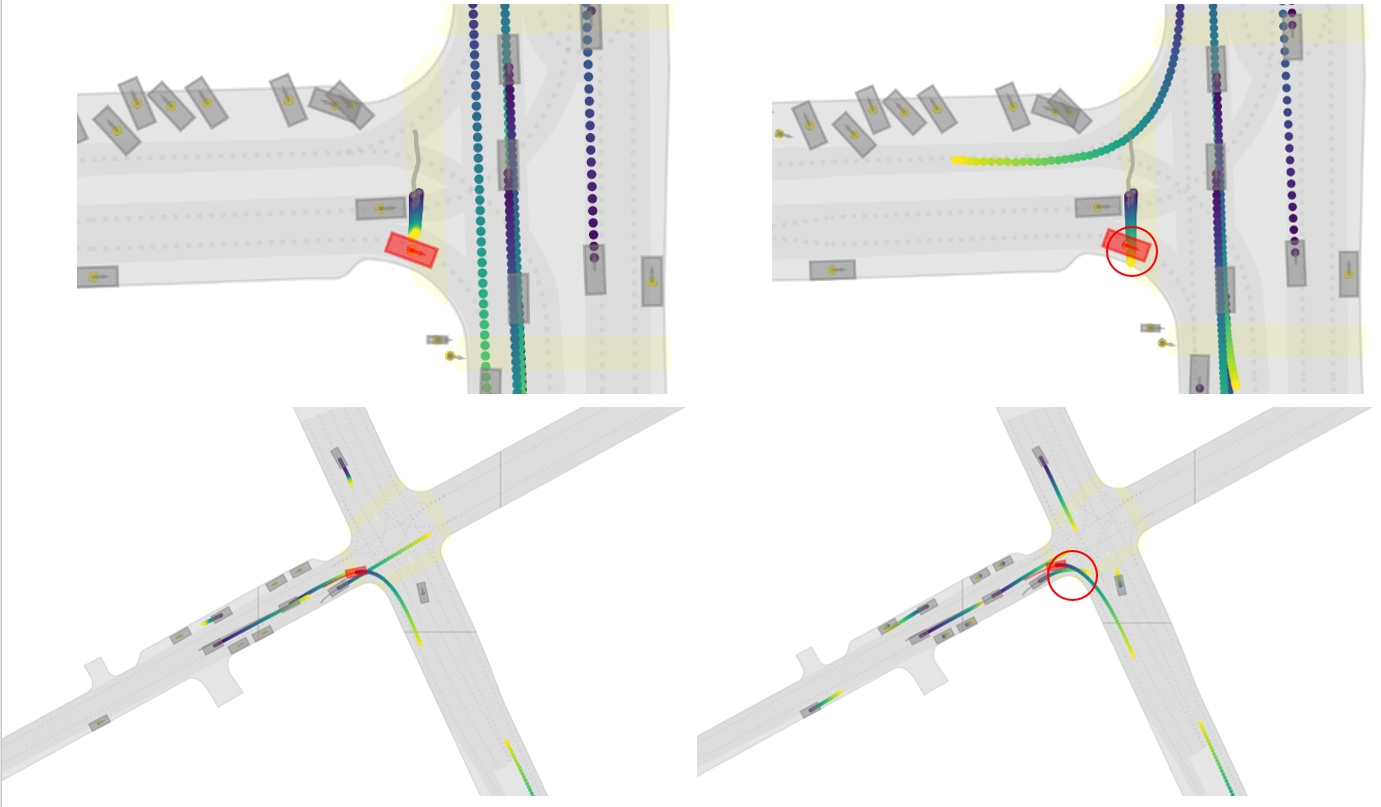}}
\caption{\textbf{Predicted Trajectories of the Two Methods} The left images shows the scene-consistent joint prediction generated using our method, while the right images displays the joint prediction produced by the straight marginal method based on the SIMPL baseline. The red circles highlight collisions that occur in the trajectories generated by the straight marginal method. In contrast, our method avoids these collisions in both scenarios, achieving better scene consistency.
}
\label{fig3}
\end{figure}

We compared the joint predicted trajectories generated by our algorithm with the baseline using three methods. The first method, Straight Marginal Prediction, assumes that the multimodal trajectories from the SIMPL baseline represent scene-level predictions. We select the scenario with the smallest Target Agent FDE\(_k\), i.e., the best predicted scenario, and calculate the FDE, ADE, MR, and CR for all jointly predicted agents in this scenario. The second method, Combined Joint Prediction, involves creating a composite trajectory by selecting the trajectory with the smallest FDE\(_k\) for each agent individually. This type of trajectory prediction is where marginal prediction excels. Finally, the joint predicted trajectories obtained using our method are referred to as Scene Joint Prediction. In this approach, we select the scenario where the trajectories of all agents are closest to the ground truth and calculate avgMinFDE\(_k\) and other metrics. These metrics directly reflects the performance of joint prediction. Fig. \ref{fig3} visualizes the joint prediction results generated by our method and by the straight marginal method for two cases. It can be observed that our method produces scene-consistent predicted trajectories and avoids collisions, whereas the straight marginal method results in trajectory collisions. The quantitative comparison of these methods is presented in the table below.
\begin{table}[htbp]
    \centering
    \setlength{\tabcolsep}{3pt}
    \caption{Quantitative results on the Argoverse 2 validation dataset}
    \begin{tabular}{lcccccc}
        \toprule
        Method & avgMinADE\(_k\)$\downarrow$ & avgMinFDE\(_k\)$\downarrow$ & avgMR\(_k\)$\downarrow$ & avgCR\(_k\)$\downarrow$\\
        \midrule
        Combined Joint & \textbf{0.48} & \textbf{0.89} & \textbf{0.10} & 0.02\\
        Straight Marginal & 0.93 & 2.44 & 0.38 & 0.02\\
        Scene Joint (Ours) & \underline{0.68} & \underline{1.52} & \underline{0.24} & \textbf{0.01}\\
        \bottomrule
    \end{tabular}
\end{table}

In the table, the best indicators are highlighted in bold font, while the second-best indicators are denoted with underlines. It is worth noting that Combined Joint has a natural advantage in the ADE, FDE, and MR metrics, as it selects the best trajectory from the \( K \) multimodal trajectories for each agent. This results in the best static imitation metrics. However, its performance in measuring trajectory collision rate and scene consistency, as indicated by the CR metric, is poor. This suggests that combined trajectories do not inherently possess scene consistency. In contrast, our method achieves the best results in these metrics, demonstrating the superiority of our modeling approach in generating scene-consistent joint predicted trajectories. Notably, its imitation metrics still outperform those obtained directly from the marginal prediction model.

\begin{table}[htbp]
    \centering
    \setlength{\tabcolsep}{3pt}
    \caption{Training and inference complexities of the three methods}
    \begin{tabular}{lcccccc}
        \toprule
        Method & Training Complexity & Inference Complexity \\
        \midrule
        Combined Joint & $O(K \times N)$ & $O(K^N)$ \\
        Straight Marginal & $O(K \times N)$ & $O(K \times N)$ \\
        Scene Joint (Ours) & $O(K \times N)$ & $O(K \times N)$ \\
        \bottomrule
    \end{tabular}
\end{table}

From the perspective of algorithmic complexity, the complexity of our algorithm in calculating the loss function is consistent with that of straight marginal prediction, both being $O(K \times N)$, where $N$ represents the number of agents to be predicted in the scene, and $K$ is the number of future time steps to be predicted. In contrast, the computational complexity of the combined joint prediction method is $O(K^N)$, which becomes highly time-consuming when $N$ is large. The computational complexities of the algorithms are shown in the table  II.

\section{Conclusion}
In this paper, we propose a method for implicit scene-consistent joint trajectory prediction using a loss function. We selected SIMPL as our baseline and enhanced it by incorporating a scene-level loss function during training. This approach resulted in joint predictions at the scene level and reduced the collision rate among the predicted trajectories of target agents within the scene. Experimental results on the Argoverse2 dataset demonstrate the effectiveness of our method. Our algorithm also has limitations. Although it is a simple and effective approach, the joint distribution of trajectories in a scene grows exponentially with the number of agents. As a result, predicting joint trajectories is more susceptible to mode collapse compared to single-agent marginal predictions, making it significantly more difficult to accurately model key agents. Meanwhile, due to the immense latent space involved in scene-level trajectory prediction, the probabilistic outcomes lose their practical significance. In the future, we intend to further investigate methods to address these challenges.

\end{document}